\documentclass[lettersize,journal]{IEEEtran}
\usepackage{amsmath,amsfonts}
\usepackage{algorithmic}
\usepackage{algorithm}
\usepackage{array}
\usepackage[caption=false,font=normalsize,labelfont=rm,textfont=rm]{subfig}
\usepackage{hyperref}
\usepackage{cleveref}
\crefname{figure}{fig}{figures}
\Crefname{figure}{Fig.}{Figures}
\usepackage{textcomp}
\usepackage{subcaption}
\usepackage{stfloats}
\usepackage{orcidlink}
\usepackage{multirow}
\usepackage{mathtools}
\makeatletter

\newcommand{\Rmnum}[1]{\expandafter\@slowromancap\romannumeral #1@}
\makeatother
\usepackage{url}
\usepackage{verbatim}
\usepackage{graphicx}
\usepackage{cite}
\hypersetup{hidelinks,
	colorlinks=true,
	allcolors=black,
	pdfstartview=Fit,
	breaklinks=true}

\begin{document}

\title{SKeDA: A Generative Watermarking Framework for Text-to-video Diffusion Models}

\author{Yang Yang~\orcidlink{0000-0003-1048-7994}, Xinze Zou~\orcidlink{0009-0003-1810-3841}, Zehua Ma~\orcidlink{0000-0002-8153-341X}, Han Fang~\orcidlink{0000-0001-9635-9859} and Weiming Zhang~\orcidlink{0000-0001-5576-6108}
\thanks{This work was supported in part by the National Natural Science Foundation of China under Grant 62272003, in part by the Quantum Science and Technology-National Science and Technology Major Project under Grant 2021ZD0302300, in part by the Science and Technology Major Project of Anhui Province under Grant 202423s06050001. \textit{(Corresponding authors: Han Fang; Zehua Ma.)}

Yang Yang and Xinze Zou are with the School of Electronic and
Information Engineering, Anhui University, Hefei, Anhui 230601, China
(e-mail: sky yang@ahu.edu.cn; p23201049@stu.ahu.edu.cn).

Zehua Ma and Weiming Zhang are with Anhui Province Key Laboratory of Digital Security and the CAS Key Laboratory of Electromagnetic Space Information, University of Science
and Technology of China, Hefei 230026, China (e-mail: mzh045@ustc.edu.cn; zhangwm@ustc.edu.cn).

Han Fang is with the School of Computing, National University of
Singapore, Singapore 117417 (e-mail: fanghan@nus.edu.sg).

}
}



\maketitle

\begin{abstract}
The rise of text-to-video generation models has raised growing concerns over content authenticity, copyright protection, and malicious misuse. Watermarking serves as an effective mechanism for regulating such AI-generated content, where high fidelity and strong robustness are particularly critical. Recent generative image watermarking methods provide a promising foundation by leveraging watermark information and pseudo-random keys to control the initial sampling noise, enabling lossless embedding. Meanwhile, the invertibility of DDIM further allows robust watermark extraction. However, directly extending these techniques to videos introduces two key limitations: 
(1) Existing designs implicitly rely on strict alignment between video frames and frame-dependent pseudo-random binary sequences used for watermark encryption. Once this alignment is disrupted, subsequent watermark extraction becomes unreliable; and (2) Video-specific distortions, such as inter-frame compression, significantly degrade watermark reliability. To address these issues, we propose \textbf{SKeDA}, a generative watermarking framework tailored for text-to-video diffusion models. \textbf{SKeDA} consists of two components: 
(1) Shuffle-Key-based Distribution-preserving Sampling (SKe) employs a single base pseudo-random binary sequence for watermark encryption and derives frame-level encryption sequences through permutation. This design transforms watermark extraction from synchronization-sensitive sequence decoding into permutation-tolerant set-level aggregation, substantially improving robustness against frame reordering and loss; and (2) Differential Attention (DA), which computes inter-frame differences and dynamically adjusts attention weights during extraction, enhancing robustness against temporal distortions.
Extensive experiments demonstrate that SKeDA preserves high video generation quality and significantly improves watermark robustness under compression, frame deletion, and noise, outperforming existing baselines in both fidelity and traceability.

\end{abstract}

\begin{IEEEkeywords}
text-to-video, copyright protection, pseudo-random key,  generative watermark,  diffusion model
\end{IEEEkeywords}

\newpage
\section{Introduction}
\IEEEPARstart{T}{he} rapid advances in generative AI, especially text-to-video diffusion models such as CogVideo~\cite{cogvideo}, Make-A-Video~\cite{makeavideo}, and Video Diffusion~\cite{videodiffusionM}, have transformed digital content creation. These systems can synthesize visually compelling and semantically coherent videos directly from natural language prompts, dramatically lowering the barrier to video production. However, the same capabilities have also intensified concerns surrounding copyright protection, provenance verification, and the large-scale dissemination of synthetic or misleading media. AI-generated videos may replicate copyrighted materials or fabricate photorealistic content, leading to risks of misattribution and misuse~\cite{tdsc1, tdsc3}. In response, regulatory frameworks such as the EU AI Act~\cite{AIACT} and the U.S. COPIED Act~\cite{COPIED ACT} increasingly mandate traceability mechanisms for synthetic media. Ensuring that generated videos carry strong, verifiable authenticity signals has thus become essential for safeguarding intellectual property and maintaining trust in digital ecosystems.

\begin{figure}[t]
    \centering
    \includegraphics[width=9cm]{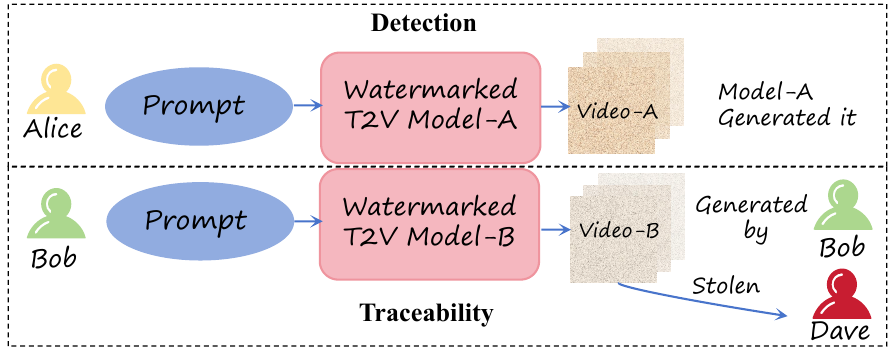}
    \caption{Application scenarios. The proposed method can detect the source of the model that generated the video, as well as track which user created the video.}
    \label{fig_1}
\end{figure}

Digital watermarking provides a principled mechanism for verifying the authenticity and provenance of generated content by embedding imperceptible signals during the generation process. As illustrated in \Cref{fig_1}, a watermark enables reliable attribution in both benign and adversarial scenarios—for example, allowing a user like Alice to prove that a video originates from Model~A, or preventing an adversary such as Dave from misappropriating Bob’s generated content. For a watermarking system to function effectively in such real-world cases, two properties are essential: fedility and robustness. Fedility requires that the embedded watermark remain imperceptible to human viewers and introduce no degradation to the visual quality of the generated video. Robustness ensures that the watermark can still be faithfully recovered after the video undergoes common transformations such as compression, resizing, noise corruption, transmission artifacts, or temporal manipulations. These two properties jointly ensure that watermarking can provide trustworthy attribution while preserving the intended viewing experience.

Traditional video watermarking methods typically applied a post-hoc embedding procedure to embed watermark signals into the pixel domain or the compressed domain. For example, compressed-domain approaches modify AC coefficients in H.264/AVC  streams~\cite{h264avc} to embed bits during encoding, while deep models such as VideoSeal~\cite{videoseal} and DVMark~\cite{dvmark} use neural encoders to inject watermark information into spatial-temporal features. Although these methods can achieve moderate robustness, they suffer from two fundamental drawbacks. First, post-hoc embedding inevitably introduces perturbations to completed video frames, often causing visual artifacts or compression-induced distortion that compromise invisibility. Second, robustness remains limited: watermark signals are fragile under strong video-specific operations such as high-rate compression, temporal cropping, frame deletion, or re-encoding. Consequently, traditional watermarking schemes struggle to simultaneously maintain imperceptibility and resilience in modern generative video pipelines.

Recent progress in generative image watermarking offers a compelling alternative by embedding watermark information directly in the latent noise of diffusion models. Techniques such as Gaussian Shading~\cite{guassianshading} achieve provably lossless visual fidelity because watermark signals influence only the initial noise distribution rather than perturbing the generated pixels. Their deterministic inversion procedures further enable reliable watermark extraction without retraining the generative model. However, extending these image-based generative watermarking techniques to video generation introduces two structural limitations. 
First, existing designs implicitly rely on strict alignment between video frames and frame-dependent pseudo-random binary sequences used to encrypt watermark bits prior to noise sampling. Common video operations such as frame deletion, reordering, or re-encoding can easily disrupt this alignment, causing bit-level desynchronization and unreliable watermark extraction.
Second, video-specific distortions, such as inter-frame compression, frame reordering, or temporal jitter, destroy the temporal alignment required for inversion-based extraction, causing the watermark signal to desynchronize across frames. These issues make naïve extensions of image watermarking unsuitable for the temporal and structural characteristics of text-to-video diffusion.

To address these challenges, we introduce SKeDA, a generative watermarking framework designed specifically for text-to-video diffusion models. SKeDA contains two complementary components. 
The Shuffle-Key distribution-preserving sampling (SKe) module generates a single base pseudo-random binary sequence for watermark encryption, while frame-level variations are introduced solely through permutation. By constraining frame-wise encryption masks to permutations of a shared base sequence, SKe decouples watermark extraction from strict frame-level alignment and transforms synchronization-sensitive sequence decoding into permutation-tolerant set-level aggregation, enabling robust watermark recovery under temporal desynchronization. Importantly, this design preserves the Gaussian noise sampling distribution required by diffusion models while maintaining sufficient randomness across frames.
The Differential Attention (DA) module enhances extraction robustness by computing inter-frame differences to identify stable regions and adaptively reweighting temporal attention during inversion, allowing the system to remain reliable under compression, frame deletion, and temporal distortions. Together, these modules embed watermark information directly within the diffusion trajectory, achieving high invisibility, temporal robustness, and scalable traceability for long videos.

The experimental results show that SKeDA achieves deep fusion of watermark and content at the latent space level of the diffusion model, taking into account invisibility, traceability and robustness. The robustness in common video compression scenarios such as H.264/H.265 is improved by 5\%-20\% on the basis of mainstream methods, while maintaining excellent performance in visual fidelity. The framework provides an efficient and robust technical solution for copyright protection and traceability of generated video content.

The contributions of this paper are as follows:
\begin{itemize}

\item We propose a novel generative video watermarking method that simultaneously preserves generation quality and enhances robustness. To maintain the native fidelity of text-to-video diffusion models, we introduce the Shuffle-Key distribution-preserving sampling (SKe) module. SKe employs a single base pseudo-random sequence and derives frame-level variations through permutation, preserving the Gaussian noise sampling distribution while enabling imperceptible latent watermark embedding and robust watermark extraction under severe temporal misalignment.
\item  We propose an inter-frame difference attention DA module for the extraction phase, which adaptively assigns the extraction weight for each frame by using the difference of each frame of the video over time. The DA module enhances the stability of watermark retrieval under various distortions (including compression, frame loss, and noise), thereby greatly improving robustness without any retraining of the text-to-video model.
\item Extensive experiments have proven that our solution is able to withstand a variety of attacks, including video compression, frame manipulation, and noise addition, while maintaining high video quality and high extraction accuracy.
\end{itemize}

\section{Related Work}
\subsection{Text to Video Generative Models}
Text-to-Video Generation \cite{cogvideo}, \cite{videofactory}, \cite{surveyai} is a cutting-edge direction in the field of generative AI, which aims to generate video content that conforms to the description based on natural language descriptions. In recent years, significant progress has been made in the development of text-to-video models based on diffusion models, which generate high-quality, time-consistent videos by modeling the video generation process as a denoising process from noise to data. The text-to-video model is usually based on the diffusion model, which extends the image diffusion model to the video generation task. CogVideo is an early open-source text-to-video model, which achieves continuity between video frames by adding Temporal Attention Modules on the basis of the pre-trained Wensheng image model. Make-A-Video proposed Pseudo-3D Convolution and Temporal Self-Attention to capture the temporal dynamics of video. Imagen Video uses Cascaded Diffusion Models to generate low- to high-resolution videos in multiple stages to improve the quality of the video. VideoCrafter \cite{videocrafter} and Lumiere \cite{space-timeDM} further optimize the temporal modeling, introduce Spatio-Temporal Decomposition and efficient sampling strategies, and significantly reduce the computational cost.

Recently, architectures based on a combination of variational autoencoder (VAE) and diffusion model have become mainstream. Sora \cite{sorafuture} is an advanced AI video generation tool that can create realistic and imaginative scenes based on text instructions. Open-source models such as Open-Sora \cite{opensora} and Latte \cite{latte} borrow from Sora's architecture to provide more efficient training and inference methods. These models typically operate in Latent Space, where the video is compressed into a low-dimensional representation by a variational autoencoder, which is then denoised by the diffusion model.

While significant progress has been made in the generation of text-to-videos, several challenges remain. First, computationally expensive and requiring a lot of GPU resources for training and inference. Efficient Video Diffusion \cite{efficiency} and FastVideo \cite{slidingtile} attempt to alleviate this problem through model compression and efficient sampling algorithms. Secondly, the temporal consistency and semantic accuracy of the generated video still need to be improved, especially in the generation of long videos. LongVideoBench \cite{longvideobench} provides a benchmark for long video generation, highlighting the limitations of existing models in complex scenarios. In addition, there is a growing concern about the copyright and ethics of generated content, and initiatives such as Hoshiar \cite{copyrightai} have examined the concept of authorship, a cornerstone of copyright, and its applicability to AI.

\subsection{Video Watermarking}
Video watermarking technology aims to achieve copyright protection, content authentication, and source tracking by embedding invisible identification information in videos \cite{guassianshading}, \cite{videoshield, revmark, tdsc2}. Traditional video watermarking technology is mainly used to protect the copyright of digital media and prevent unauthorized copying. Asikuzzaman and Pickering \cite{overviewvwm} provide a comprehensive overview of embedding methods based on domains such as discrete cosine transform (DCT) \cite{dct}, discrete wavelet transform (DWT) \cite{dwt}, and singular value decomposition (SVD) \cite{svd}. These methods typically embed watermark information into the Frequency Domain or Spatial Domain of the video to ensure that the watermark is robust to attacks such as compression, cropping, and noise. For example, Cox et al. \cite{digtalwm} proposed a DCT-based watermarking scheme to achieve a balance between concealment and robustness by embedding watermarks in the low-frequency components of the video frame. Hartung and Girod \cite{digitalcompress} developed a watermarking method based on the MPEG compression domain for real-time video streaming.

In recent years, deep learning technology has been introduced into traditional video watermarking. Zhang et al. \cite{robustinvisible} proposed a robust watermarking method based on attention mechanism to enhance resistance to geometric attacks by embedding watermarks in high-information areas in video frames. Fernandez et al. \cite{videoseal} developed a comprehensive framework for neural video watermarking and a competitive open-sourced model. These methods significantly improve the robustness of watermarks, but they still focus on traditional video content rather than the dynamic nature of AI-generated videos.

The limitation of traditional video watermarking is that it is often embedded in a post-hoc processing manner, which can lead to degraded quality or insufficient adaptability to complex transformations. In addition, traditional methods often assume that video content is static and difficult to adapt to the generation process of text-to-video models, which prompted researchers to turn to new methods of embedding watermarks in the generation process.
The diffusion model is the mainstream architecture of text-to-video generation, so many methods for generating video watermarks are directly designed for the diffusion model. VideoShield \cite{videoshield} is a watermarking framework specifically designed for diffusion video generation models, which embeds watermarks directly during the generation process by mapping watermark bits to template bits and generating watermark noise during denoising. This method does not require additional training and supports tamper detection in time and space, which is highly relevant to our method in terms of embedding timing. DiffuseTrace \cite{diffusetrace} proposes a transparent watermarking scheme, which unifies the watermark information with the initial latent variable through the encoder-decoder model and embeds it into the sampling process of the diffusion model. The proposed method shows strong resistance to attacks based on Variational Autoencoder (VAE) and diffusion model.

Latent space watermark is another important direction to generate video watermarks. Fernandez et al. \cite{stablesignature} proposed a Stable Signature method to achieve stability in model updates by embedding binary signatures during the sampling process of latent diffusion models. Wen et al. \cite{treering} proposed Tree-ring Watermarks, which generate invisible and robust watermarks by embedding specific patterns in the latent space of the diffusion model. These methods are primarily aimed at image generation, but their ideas can be extended to video generation. For example, Dvmark proposed by Luo et al. \cite{dvmark} is a deep learning-based video watermarking framework that enhances robustness to video editing and compression by embedding watermarks through multi-scale features, but in some cases, watermark embedding may have a slight impact on the visual quality of the video.

\begin{figure*}[t]
    \centering
    \includegraphics[width=1\textwidth]{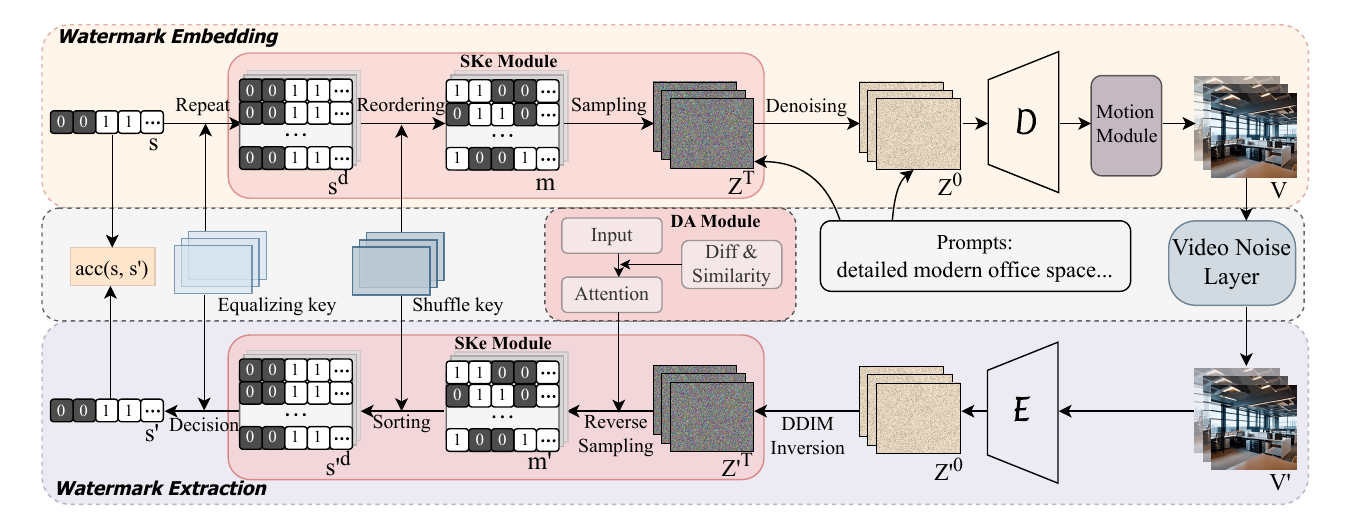}
    \caption{The framework of SKeDA. Our method consists of four main components: encryption, embedding, extraction, and decryption. In the encryption and embedding stages, the SKe module uses random shuffle key to distribute and rearrange in the latent space to realize the hidden embedding of watermark information without affecting the video quality. In the extraction and decryption stage, the DA module adaptively assigns weights based on the inter-frame difference in the extraction stage, to improve the robustness and retrieval accuracy of the watermark under various distortion conditions.}
    \label{framework}
\end{figure*}

\section{Proposed Method}
In this section, we elaborate on our proposed watermarking scheme for text-generated videos. This method embeds a watermark into the latent space of a diffusion model prior to video generation, ensuring seamless integration with the content. The approach is designed to achieve robustness against various attacks, maintain high visual fidelity, and eliminate the need for retraining the underlying text-to-video model. The framework of our method as shown in \Cref{framework}. They are respectively the watermark embedding and watermark extraction stages, each comprising distinct steps detailed below.

\subsection{Watermark Embedding}
\subsubsection{\bf{Message Encoding}}
In the initial stage of watermark embedding, we embed the watermark information seamlessly into the latent space of the video while ensuring imperceptibility and robustness. The watermark is represented as a binary string within the set $\{0, 1\}$. This binary string can encode various identifiers, such as copyright notices, authentication tokens, or unique ownership tags, which are customized for specific use cases. And the initial size is defined as $\left(\frac{f}{k_f}, \frac{c}{f_c}, \frac{h}{f_h}, \frac{w}{f_w}\right)$ bits. Among them, $f$ represents the feature dimension, $c$ represents the number of channels, $h$ and $w$ correspond to the height and width of the latent space, respectively. The scaling factors $k_f$, $f_c$, $f_h$ and $f_w$ are the corresponding hyperparameters, which are used to adjust the size of the watermark to be consistent with the structure of the latent space $(f, c, h, w)$. Then we use a randomly generated equalizing key of 0,1 uniformly distributed to convert the watermark information into a uniformly distributed sequence through XOR operation, to prevent the impact of uneven distribution of watermark information on the embedding process. The extended process involves copying the watermark bits into the target dimension, and make it adapt to the input size of the model while maintaining the integrity of the original information. This step can ensure that the watermark is evenly distributed throughout the latent representation, and enhance its ability to resist local distortion or attack.
\begin{equation}\label{reorder}
    m=s^{d}\left[k_{i}\right] \mid k_{i} \in k e y, i=1,2, \ldots, n
\end{equation}

\subsubsection{\bf{SKe Module}}
Next, we introduce the SKe module, which includes scrambling operations for extended secret messages and sampling of encrypted messages, using random keys to ensure lossless visual quality, and embedding watermark information into each video frame. First, we employ an Equalizing key to uniformly distribute the original message, ensuring an approximately uniform statistical distribution. Then, we use a randomly shuffled sequence as the encryption key, this key enables the reordering watermark information to maintain a high degree of randomness while maintaining an even distribution, which is convenient for subsequent sampling to conform to the Gaussian distribution and achieve lossless generation quality. Furthermore, this design minimizes the dependence on multiple keys while preserving the visual fidelity of the generated frames, thereby enabling generation for long video sequences. The encryption process is performed by reordering the elements of the secret information $s^d$ using the generated key $key$, which generates an encrypted message $m$, as shown in Eq.~(\ref{reorder}). This encryption method ensures that the embedded information remains hidden even if the hidden space is intercepted or analyzed.

To map the encrypted message $m$ into the latent space,
as shown in \Cref{framework}, for any bit $i\in\{0,1\}$ in message $m$, which corresponds to the $\alpha$ in $Z^T$, here we use $f(\cdot)$, $ppf(\cdot)$ to represent the probability density function and percentile function of the Gaussian distribution $\mathcal{N}(0,1)$, then we have:

\begin{equation}\label{eq1}
p(\alpha|i)=\begin{cases}
2f(\alpha),&ppf(\frac{i}{2})<\alpha\leq ppf(\frac{i+1}{2}) \cr
0,&otherwise \end{cases}.\\
\end{equation}  
At this point, $Z^T$ is also a standard Gaussian distribution, and we can get the probability distribution of $\alpha$ as: 
\begin{equation}\label{eq2}
    p(\alpha)=\sum_{i=0}^1p(\alpha|i)p(i)=\frac{1}{2}(p(\alpha|0)+p(\alpha|1))=f(\alpha).
\end{equation}  
After that, we perform iterative denoising on noise state $Z^T$ to obtain the latent representation $Z^0$ after denoising. Here we adopt the DPMSolver \cite{dpmsolver} algorithm, which can accelerate convergence while maintaining or enhancing the quality of the generated output. Due to its ability to optimize step size and sampling trajectory, it is very suitable for real-time or resource-constrained applications. Text embedding is used to adjust the direction of each denoising step, ensuring that the generated image conforms to the text description.

\subsubsection{\bf{Video Generation}}
After denoising, we utilize the Stable Diffusion (SD) decoder \cite{sd} and AnimateDiff's Motion Module \cite{animatediff} to convert the latent representation $Z^0$ into coherent video. The SD decoder maps spatial latent features to pixel space, generating high-quality images that retain semantic and structural information. However, as SD processes frames independently, the resulting sequence lacks temporal consistency. To address this, AnimateDiff's motion module incorporates a temporal attention mechanism and trajectory-consistent transformations to ensure inter-frame coherence. This two-stage pipeline completes the process from latent space to image decoding and then to motion-aware video synthesis, ultimately resulting in effective watermark embedding in the latent space while maintaining high visual quality and temporal smoothness for text-to-video generation.


\subsection{Watermark Extraction}
\subsubsection{\bf{DDIM Inversion}} In the watermark extraction stage, we begin by processing the attacked watermarked video $V'$ to recover the embedded watermark. The first step is to encode $V'$ back into the latent space using the Stable Diffusion Encoder, which maps the high-dimensional visual content into a compact latent representation $Z^{'0}$. This encoding process effectively inverts the original decoding path and enables us to operate directly in the latent domain, where the watermark was originally embedded.

To retrieve the latent watermark information, we further reconstruct the noisy latent representation $Z^{'T}$ corresponding to the initial stage of the forward diffusion process. This is achieved using the DDIM (Denoising Diffusion Implicit Models) inversion technique \cite{denoising}, which simulates the forward diffusion steps in a deterministic and controllable manner. By progressively adding noise to $Z^{'0}$ over a predefined number of time steps, the DDIM inversion produces a noisy implicit representation of $Z^{'T}$ that approximates the initial Gaussian noise state.

This inversion is crucial for watermark recovery, as it restores the latent conditions under which the watermark was originally embedded. Unlike stochastic diffusion models, DDIM provides a deterministic inversion path, which enhances consistency and fidelity in watermark extraction. Moreover, by operating entirely in the latent space, our approach remains robust to a wide range of distortions and video-level attacks, ensuring that the embedded watermark features are preserved and remain recoverable under adverse conditions.



\subsubsection{\bf{Watermark Decoding based on DA Module}} In the decoding phase, our goal is to recover the watermark message $s'$ from the noisy latent representation $Z^{'T}$. Firstly, the embedded bit $m'$ is reconstructed by reverse sampling and analyzing the noisy latent representation. In order to improve the extraction accuracy, especially in the case of distortion, we have added DA module. The module utilizes the temporal information in the video by evaluating the similarity between the first frame and the rest of the video (cosine distance), dynamically assigning weights to each frame. Frames with high similarity have a lower degree of variation in motion and are given greater weights to amplify their contribution to the watermark recovery process, while frames with low similarity are given less weight to mitigate the potential impact. The cosine similarity $S_t$ between the first frame $f_1$ and the $t$ frame $f_t$ is calculated as follows:
\begin{equation}\label{cosine}
    S_{t}=\frac{f_{1} \cdot f_{t}}{\left\|f_{1}\right\|\left\|f_{t}\right\|}
\end{equation}
Then, the attention score $A_{j, i}$ is calculated by the similarity between $j$ frame and $i$ frame based on \cite{attention}. 
Finally, the final weight $w_t$ is calculated by combining $S_t$ and $A_{t,i}$ as follows:

\begin{equation}\label{weight}
w_t=\frac{\exp( S_t+\sum_{i=1}^TA_{t,i})}{\sum_{j=1}^T\exp( S_j+\sum_{i=1}^TA_{j,i})}\end{equation}
After that, the weighted frame contributions are aggregated to calculate the encrypted message $m'$, where $m_t$ is the encrypted message obtained by reverse sampling the $t$ frame:
\begin{equation}\label{m'}
m'=\sum_{t=1}^j{w_t}\times m_t
\end{equation}

Then $m'$ is decrypted using the same shuffle key used during embedding.
Based on the information of keys, the elements of $m'$ are reversed to restore it to the state $s{'^d}$ where the secret information is repeatedly aggregated. After that, $s{'^d}$ is aggregated after being divided into ${k_f}\times {f_c\times f_h}\times {f_w}$ blocks by dimension, which are the scaling factors in the previous encoding process.
\begin{equation}\label{eq3}
    M=\sum_{k=1}^N\mathrm{s^{'d}}_k,\quad N={k_f}\times {f_c\times f_h}\times {f_w}
\end{equation}
Here, $M$ denotes the aggregate score for each block, the threshold (set at 0.5) determines the equalizing message. The final binary watermark message $s'$ is obtained by performing a XOR operation with the equalizing key:
\begin{equation}\label{eq4}
\mathrm{s'}=\begin{cases}1,&\ M>0.5\\0,&\ M\leq0.5\end{cases}
\end{equation}
This chunked aggregation improves robustness and allows watermarks to be reconstructed even if parts of the video are lost or distorted, as redundant embedding across frames and spatial regions ensures sufficient data for accurate recovery.



\section{Experiments}

\subsection{Implementation Details}
All experiments were conducted on an NVIDIA RTX 4090 GPU equipped with 24GB of virtual memory, using its high computing power to handle the intensive needs of video watermarking and diffusion-based generation. We used a combination of Stable Diffusion version 1.5 and Animatediff's Motion Module to implement our watermarking scheme. The SD model is the basis for generating video frames based on latent representations, while the motion module ensures temporal consistency between frames. Here we used the same parameter settings as most video watermarking methods such as Gaussian Shading \cite{guassianshading}. The final video output is $3 \times 16\times512\times512$, 2 seconds long, a total of 16 frames, and each frame has watermark information embedded. The latent space dimension of the embedded watermark is $16 \times 1\times4 \times 64 \times 64$. During the inference process, we configured the diffusion model into 25 time steps and 8 guidance scales to balance the generation quality and watermark embedding efficiency. During watermark detection, we evaluate performance by calculating the true positive rate (TPR) at a fixed false positive rate (FPR) $(1 \times 10^{-6})$, and at the same time calculate bit precision to measure the accuracy of hidden information extraction. 

For evaluation, we randomly sampled a subset of videos from the WebVid-10M dataset \cite{frozenintime}, a large-scale video-text corpus. We calculated the evaluation metrics – Fréchette Video Distance (FVD) \cite{fvd}, CLIP score \cite{transferablevisual}, and video quality \cite{vbench} for 10 batches of data (each containing 100 videos) to ensure statistical robustness.

\begin{figure}[t]
\centering
\subfloat[]{\includegraphics[width=0.2\linewidth]{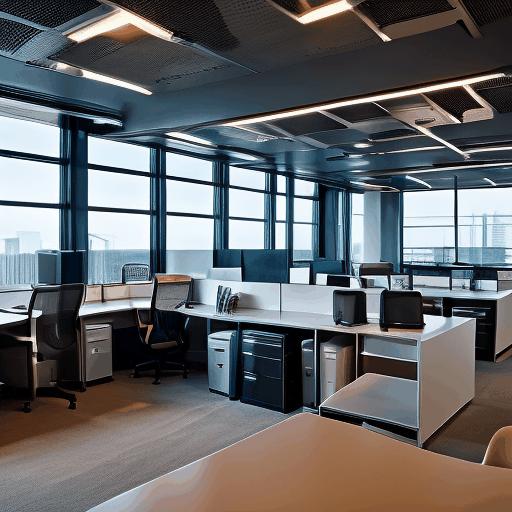}} \hspace{26pt}
\subfloat[]{\includegraphics[width=0.2\linewidth]{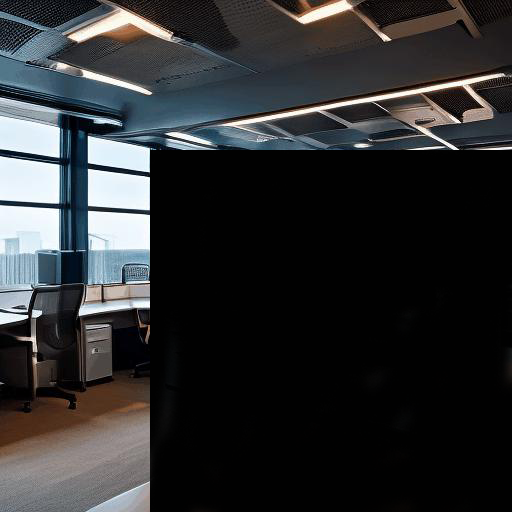}} \hspace{26pt}
\subfloat[]{\includegraphics[width=0.2\linewidth]{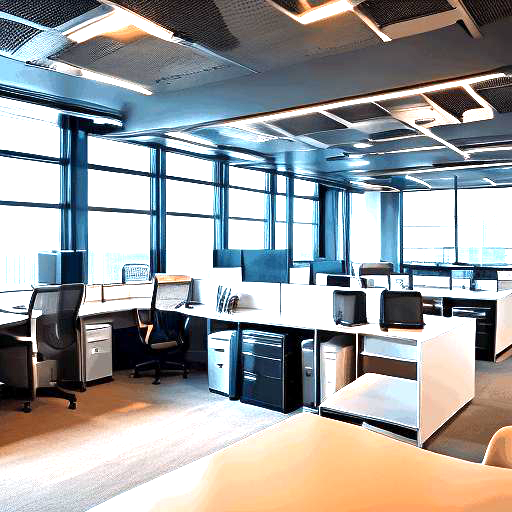}} \\[-9pt]
\subfloat[]{\includegraphics[width=0.2\linewidth]{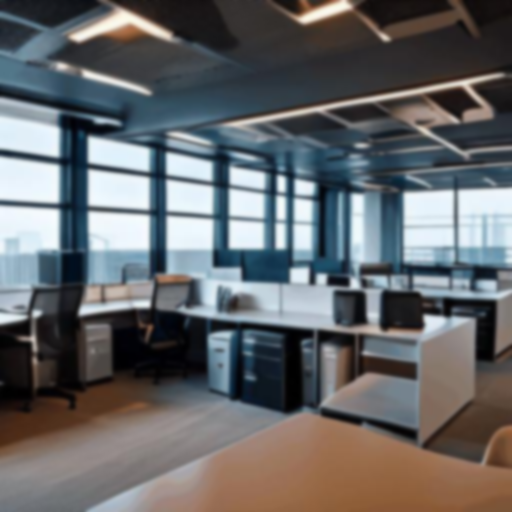}} \hspace{26pt}
\subfloat[]{\includegraphics[width=0.2\linewidth]{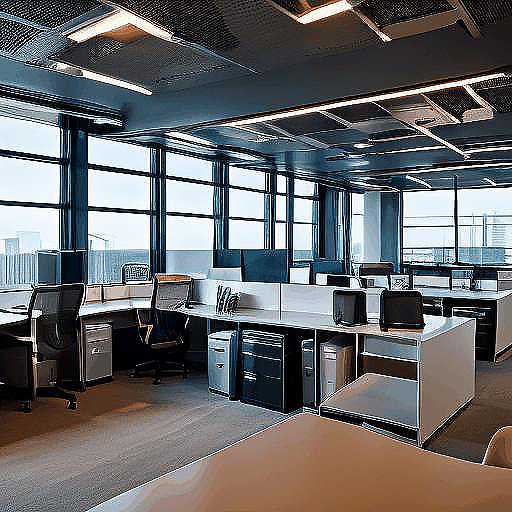}} \hspace{26pt}
\subfloat[]{\includegraphics[width=0.2\linewidth]{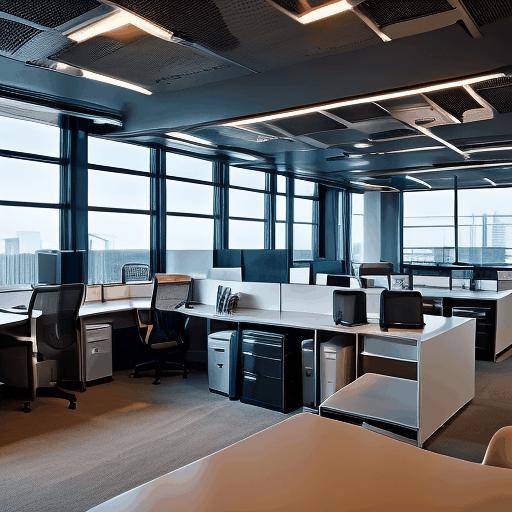}}


\caption{One frame in the video is attacked by different noises. (a) Watermarked frame. (b) 50\% Random Crop. (c) Brightness, factor=4. (d) Gaussian Blur, std=2.0. (e) Gaussian Noise, std=0.04. (f) H.264, CRF=30.}
\label{various distortion}
\end{figure}

\subsection{Comparative Experiments}
We selected HiDDeN \cite{hidden}, REVMark \cite{revmark}, Video Seal \cite{videoseal}, WAM \cite{WAM} and DVMark \cite{dvmark} as baselines. HiDDeN is a representative image watermarking model from recent years. It achieves good concealment and robustness while maintaining high payload. When adapted for video, it typically embeds the watermark frame-by-frame.
VideoSeal, developed by Meta, is an efficient and open-source video watermarking framework known for its comprehensive performance.
REVMark, Watermarking Anything Model (WAM) and DVMark are all models with good performance in recent years. Image watermarking methods typically employ a frame-by-frame embedding approach, embedding the watermark into the entire video, then extracting it frame by frame, averaging the weights of each frame, and finally fusing them together to form the final watermark information. Video watermarking methods, on the other hand, directly embed and extract watermarks from the video's multi-degree features or latent space.

{\renewcommand{\arraystretch}{1.2}
\begin{table}[tbp!]
\centering
\caption{Quality metrics for different video watermarking methods.}
\label{TABLE{quality}}
\begin{tabular}{c|ccc}
\hline
Methods    & FVD↓           & CLIP-score↑     & Video Quality↑  \\ \hline
HiDDeN\cite{hidden}     & 436.1          & 0.3077          & 0.7436          \\
REVMark\cite{revmark}    & 369.3          & 0.3287          & 0.7884          \\
Video Seal\cite{videoseal} & 365.1          & 0.3302          & 0.7799          \\
WAM\cite{WAM}        & 379.2          & 0.3146          & 0.7846          \\
DVMark\cite{dvmark}     & 382.8          & 0.3274          & 0.7742          \\
ours       & \textbf{361.3} & \textbf{0.3345} & \textbf{0.7898} \\ \hline
\end{tabular}
\end{table}}

\subsubsection{\textbf{Visual Quality Comparison}}
To measure the impact of watermark embedding on video quality, we use FVD, CLIP-score, and Video Quality indicators to test, as reported in 
Table \ref{TABLE{quality}}. In order to calculate the FVD, we use 1000 videos from the dataset Webvid-10M. The CLIP score measures the cosine similarity between each frame of the generated video and its prompt in the coding space of the CLIP model. We use VBench for video quality. It is a comprehensive benchmark test suite for video generation models, which can decompose ``video generation quality" into multiple well-defined dimensions to promote fine-grained and objective evaluation.

\textbf{Fr\'echet Video Distance (FVD)}: This metric measures the similarity between the distributions of generated and real videos, with lower values indicating better quality. Our method achieves an FVD of 361.3, outperforming baselines such as DVMark (382.8) and hidden (436.1), suggesting minimal impact on video quality.

\textbf{CLIP-score}: This evaluates the semantic alignment between the video content and the text prompt. Our method scores 0.3345, higher than baselines like REVMark (0.3287) and WAM (0.3146), indicating strong consistency with the input prompt.

\textbf{Perceptual Video Fidelity}: Assessed using VBench, it includes dimensions such as image quality, temporal consistency, and aesthetic quality. All metrics were positive, i.e., higher scores indicated better video quality in that dimension, and finally a composite score was taken. Our method achieved a score of 0.7898, surpassing all baselines, including REVMark (0.7834) and DVMark (0.7742), demonstrating a higher perceptual quality.

{\renewcommand{\arraystretch}{1.2}
\begin{table*}[t]
\centering
\caption{ Comparison results. We control the FPR at $10^{-6}$, and evaluate the bit accuracy for SD V1.5.}
\label{TABLE{comparison}}
\begin{tabular}{c|cccccccc}
\hline
Methods    & \begin{tabular}[c]{@{}c@{}}Frame Average \\ ($N$=3)\end{tabular} & \begin{tabular}[c]{@{}c@{}}Frame Drop\\ ($p$=0.5)\end{tabular} & \begin{tabular}[c]{@{}c@{}}Frame Swap\\ ($p$=0.5)\end{tabular} & \begin{tabular}[c]{@{}c@{}}Random Crop\\  ($p$=0.7)\end{tabular} & \begin{tabular}[c]{@{}c@{}}Gaussian Noise\\ (std=0.04)\end{tabular} & \begin{tabular}[c]{@{}c@{}}Gaussian Blur\\  (std=2.0)\end{tabular} & \begin{tabular}[c]{@{}c@{}}H.264\\   (CRF=30)\end{tabular} & Average \\ \hline
HiDDeN\cite{hidden}     & 0.9691          & 0.9903          & 0.9910          & 0.7727          & 0.9127          & 0.7270          & 0.7008          & 0.8662   \\
REVMark\cite{revmark}    & \textbf{0.9998}         & 0.9981        & 0.9998                 & 0.9631        & \textbf{0.9997}        & 0.9982        & 0.8577          & 0.9738   \\
Video Seal\cite{videoseal} & 0.9931                                  & 0.9919                   & 0.9931       & 0.9356          & 0.9887          & 0.9904          & 0.8556          & 0.9632   \\
WAM\cite{WAM}      & 0.9864                     & 0.9903           & 0.9908             & \textbf{0.9912}                  & 0.9822             & \textbf{0.9998}          & 0.8612          & 0.9717   \\
DVMark\cite{dvmark}    & 0.9810                & 0.9899              & 0.9935                & 0.9706               & 0.9901            & 0.9809                  & 0.7996               & 0.9579   \\
Ours       & 0.9888                     & \textbf{0.9990}                    & \textbf{0.9998}                 & 0.9585                  & 0.9890                         & 0.9932                        & \textbf{0.9687}                        & \textbf{0.9853}   \\ \hline
\end{tabular}
\end{table*}}

\begin{figure*}[tbp]
\centering
\subfloat[Frame Drop]
{
    \begin{minipage}[t]{.3\linewidth}
        \centering
        \includegraphics[width=2in]{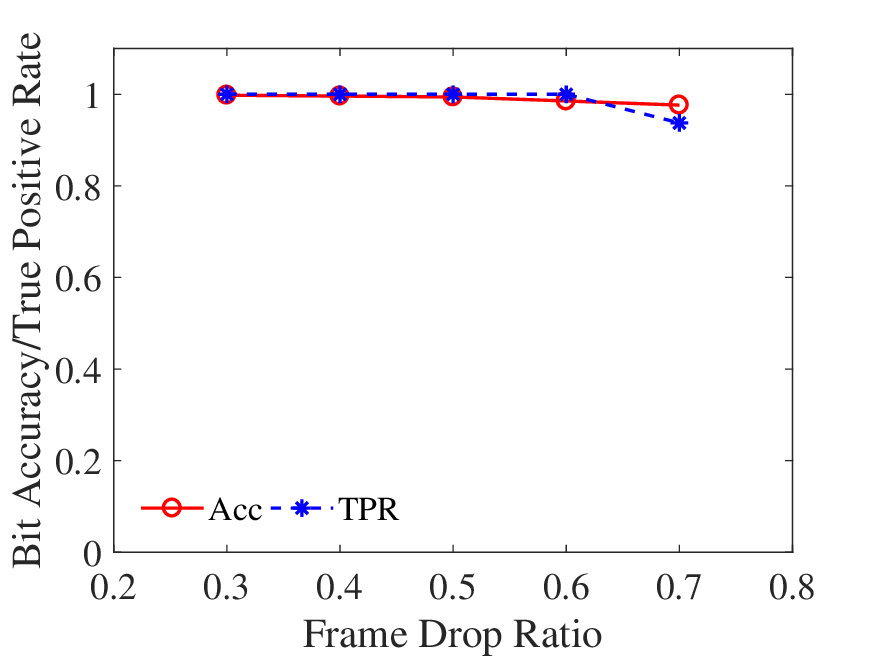}
    \end{minipage}
}\vspace{-4pt}\hspace{-8pt}
\subfloat[Crop]
{
    \begin{minipage}[t]{.3\linewidth}
        \centering
        \includegraphics[width=2in]{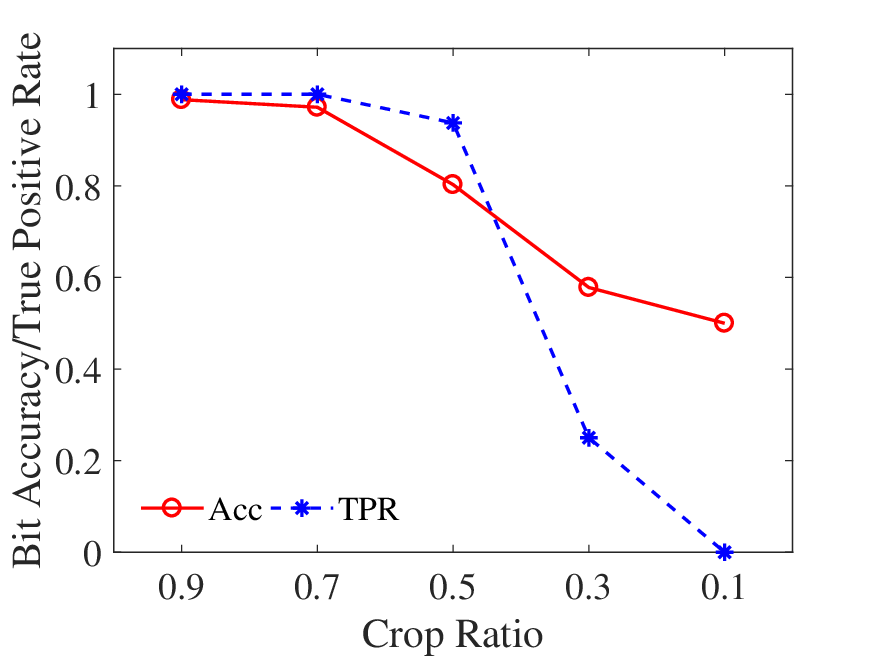} 
    \end{minipage}\label{crop}
}\vspace{-4pt}\hspace{-8pt}
\subfloat[Frame Swap]
{
    \begin{minipage}[t]{.3\linewidth}
        \centering
        \includegraphics[width=2in]{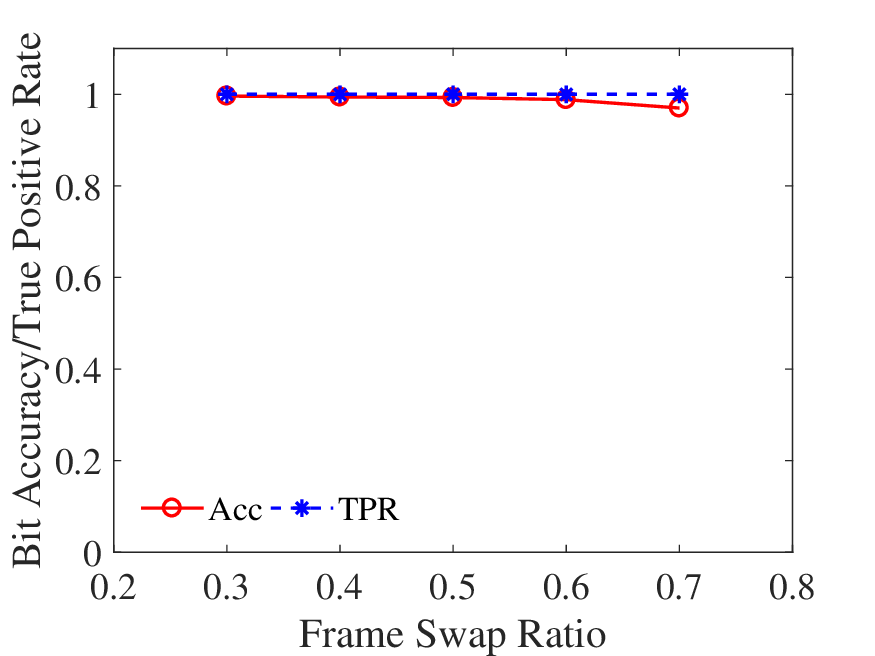} 
    \end{minipage}
}\vspace{-1pt}\hspace{-8pt}

\subfloat[Frame Average]
{
    \begin{minipage}[t]{.3\linewidth}
        \centering
        \includegraphics[width=2in]{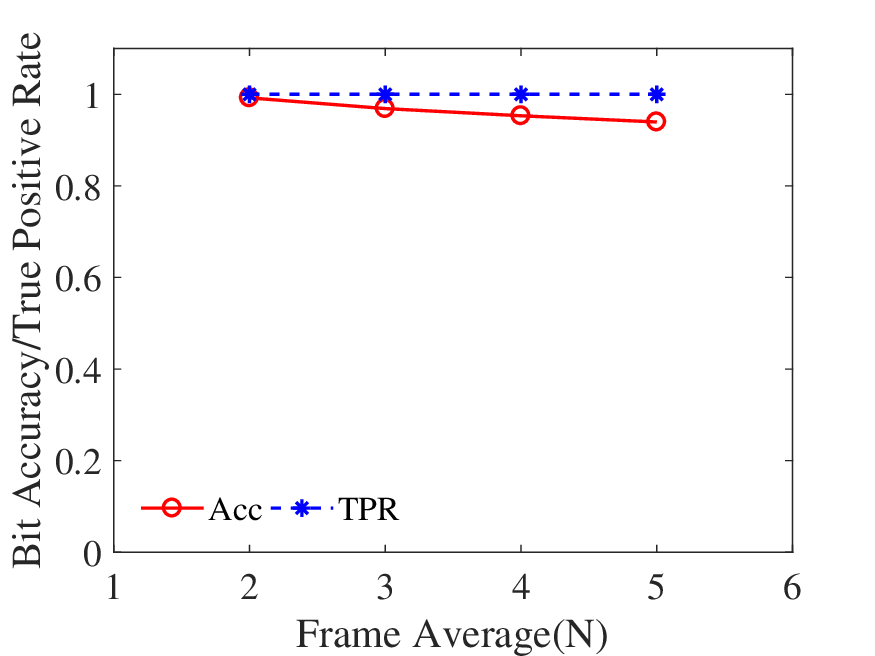} 
    \end{minipage}
}\vspace{-1pt}\hspace{-8pt}
\subfloat[Brightness]
{
    \begin{minipage}[t]{.3\linewidth}
        \centering
        \includegraphics[width=2in]{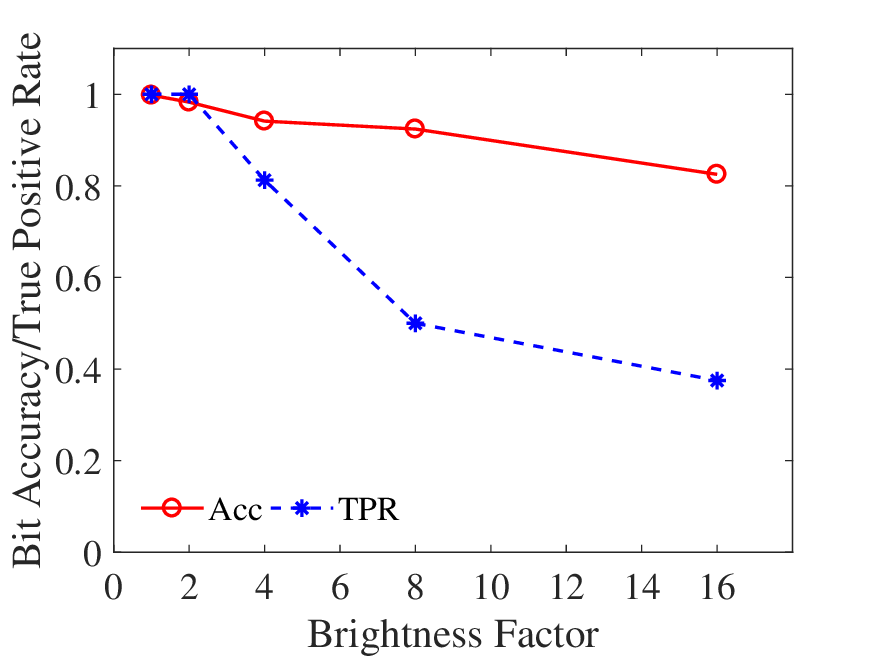}
    \end{minipage}
}\vspace{-1pt}\hspace{-8pt}
\subfloat[H.264]
{
    \begin{minipage}[t]{.3\linewidth}
        \centering
        \includegraphics[width=2in]{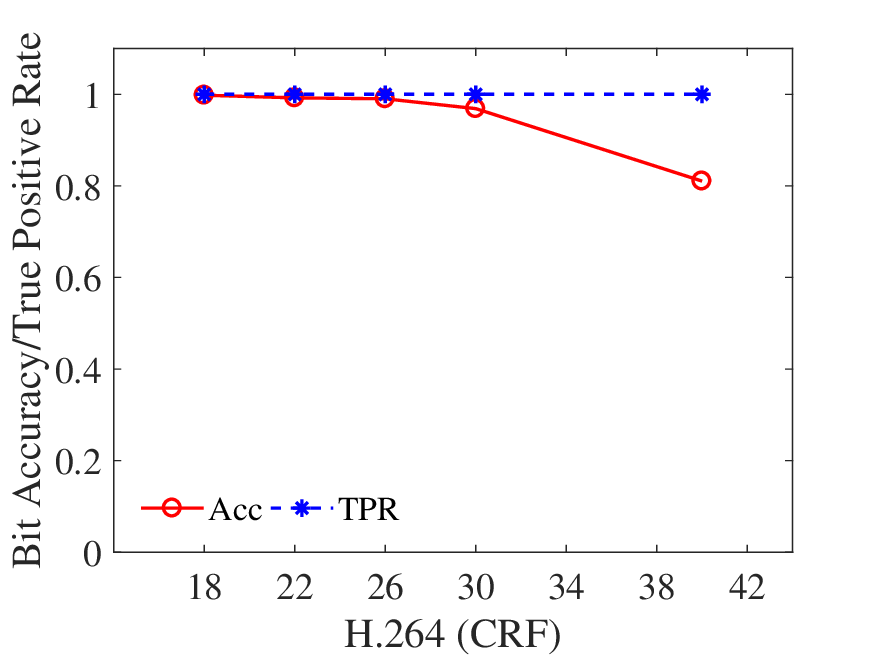} 
    \end{minipage}
}\vspace{-1pt}\hspace{-8pt}

\caption{TPR and bit accuracy under various video distortions and distortion strengths.}
\label{video_acc}
\end{figure*}

\subsubsection{\textbf{Robustness Comparison}}
To assess robustness, we measured bit accuracy, given by the percentage of bits correctly decoded. As a result, we report bitrate accuracy for each model on a large number of common distortions, including H.264, random frame drops, and averaged frames. Here we have the largest payload, WAM (32 bits), ours 256 bits, and the others are all 96 bits. We explain each type of distortion and its respective parameters in detail. H.264 refers to standard H.264 video compression with a fixed constant rate factor (CRF). In frame averaging, we average all frames within a time window, where $N$ refers to the size of the window. For example, for $N=3$, the warp frame for time $T$ is Avg$(F_{T-1},\ F_T, \ F_{T+1})$. For dropped frames and frame swaps, $p$ refers to the probability of dropped frames (or swapping with their immediate neighbors). Random cropping refers to the random cropping of a common rectangular area (for all frames) in the case of a given video. $p$ refers to the width and height ratio of the cropped frame. For example, a cropped video with $p=0.7$ would have $h' = h\times0.7$ and $w' = w\times0.7$. The standard deviation in Gaussian fuzzy and Gaussian noise refers to the standard deviation of the Gaussian kernel and the standard deviation of the random normal distribution that produces the noise. An example of a warped frame is shown in the image.

As shown in Table \ref{TABLE{comparison}}, we tested 1000 videos for each method, and our method performed well on almost all tested distortions, with an average bit accuracy improvement of 1.15\% over the best-performing baseline, especially on H.264 compression. Therefore, we further conducted comparative experiments with different compression ratios on H.264 compression, which we will explain in detail in later sections. As can be seen in Table \ref{TABLE{comparison}}, our method can still obtain good extraction accuracy under CRF=30, 10.75\% higher than the next best method; The video size at CRF=30 is only 8.3\% of CRF=0.
 This can be attributed to the widespread diffusion of watermarks throughout the latent space.

\subsubsection{\textbf{Compression Resistance Comparison}}
We evaluated the robustness of our method against video compression using two widely adopted codecs, H.264 and H.265. As shown in Tables \ref{TABLE{H264}} and \ref{TABLE{H265}}, we measured watermark extraction accuracy across multiple constant rate factor (CRF) settings, which control compression strength and file size.

Under H.264 compression, the visual quality loss is obvious as the CRF increases, and our method still performs well. For example, at CRF=30 in H.264, our method obtained a bit accuracy of 96.87\%, which is significantly better than the baseline method, such as DVMark (86.92\%) and REVMark (85.77\%). 
For H.265 compression, which achieves more aggressive bitrate reduction at the same CRF, our approach demonstrates similar resilience. While at CRF=30 in H.265, we still obtained an accuracy of 95\%, which is 7.66\% higher than other methods.

When CRF=40, the video size is only about 1\% of the original size, and our method still achieves an extraction accuracy of about 80\%. These results confirm that SkeDA can accurately extract watermark information even under high compression.

\subsection{Performance under Various Distortions}
\subsubsection{\textbf{Video-level Distortion Impact}}
To evaluate the robustness of our proposed watermarking scheme, we conducted extensive experiments to assess its performance under various common attacks and distortions that videos and images may encounter in real-world scenarios. These include video compression, frame manipulations, spatial operations, and noise addition. The objective is to ensure that the embedded watermark remains detectable even after significant alterations to the video content. Additionally, we assessed the imperceptibility of the watermark to confirm that it does not compromise the visual quality of the generated videos.

We first tested the extraction accuracy and TPR of watermarks under video attacks. As shown in \Cref{video_acc}, we can see that except for \Cref{crop} Crop, our methods achieved good results in relatively serious cases. And it can be seen that our method performs well in terms of frame dropping, frame swapping and frame averaging. This can be attributed to the fact that we have embedded watermarks in each frame, making it effectively resistant to frame attack operations.
\begin{figure*}[t]
\centering
\subfloat[JPEG]
{
    \begin{minipage}[b]{.2\linewidth}
        \centering
        \includegraphics[width=1.7in]{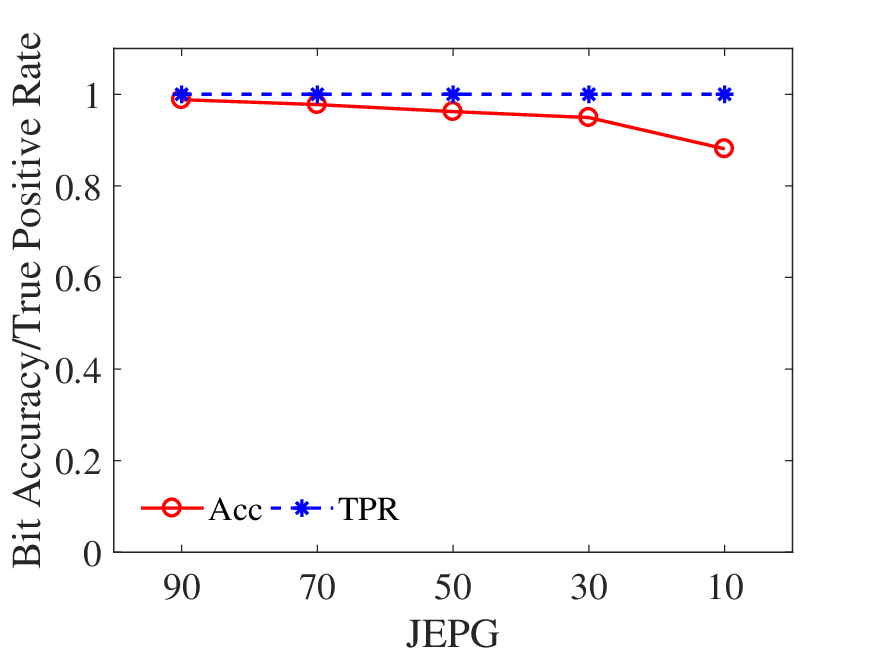}
    \end{minipage}
}\vspace{-2pt}\hspace{1pt}
\subfloat[Resize]
{
    \begin{minipage}[b]{.2\linewidth}
        \centering
        \includegraphics[width=1.7in]{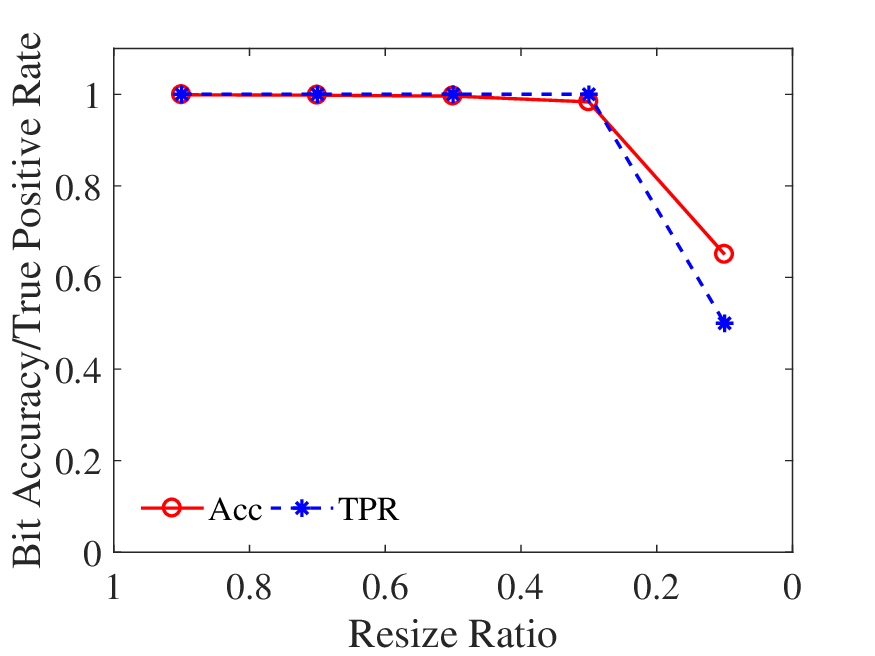} 
    \end{minipage}
}\vspace{-2pt}\hspace{1pt}
\subfloat[Median]
{
    \begin{minipage}[b]{.2\linewidth}
        \centering
        \includegraphics[width=1.7in]{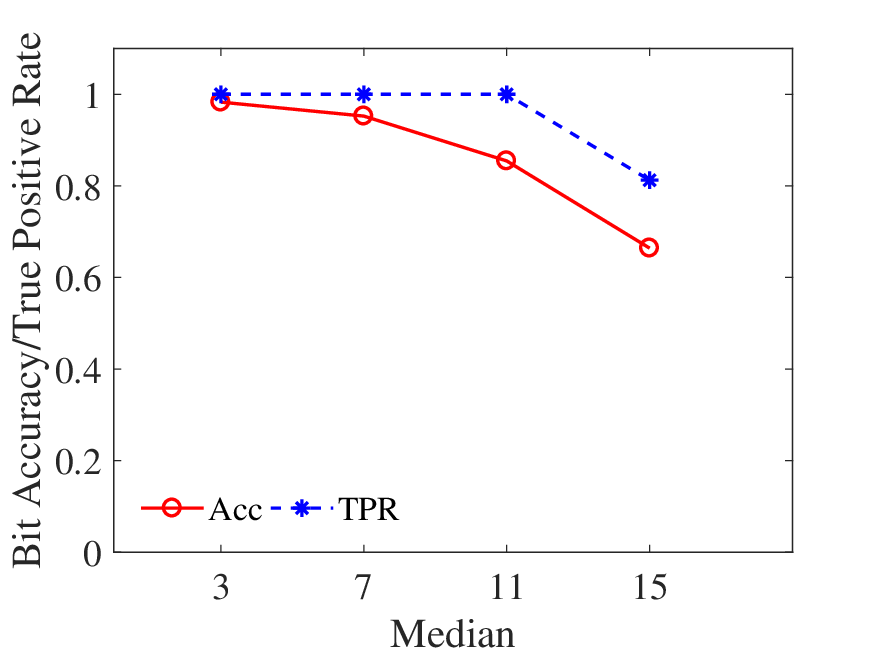}
    \end{minipage}
}\vspace{-2pt}\hspace{1pt}
\subfloat[Salt Pepper]
{
    \begin{minipage}[b]{.2\linewidth}
        \centering
        \includegraphics[width=1.7in]{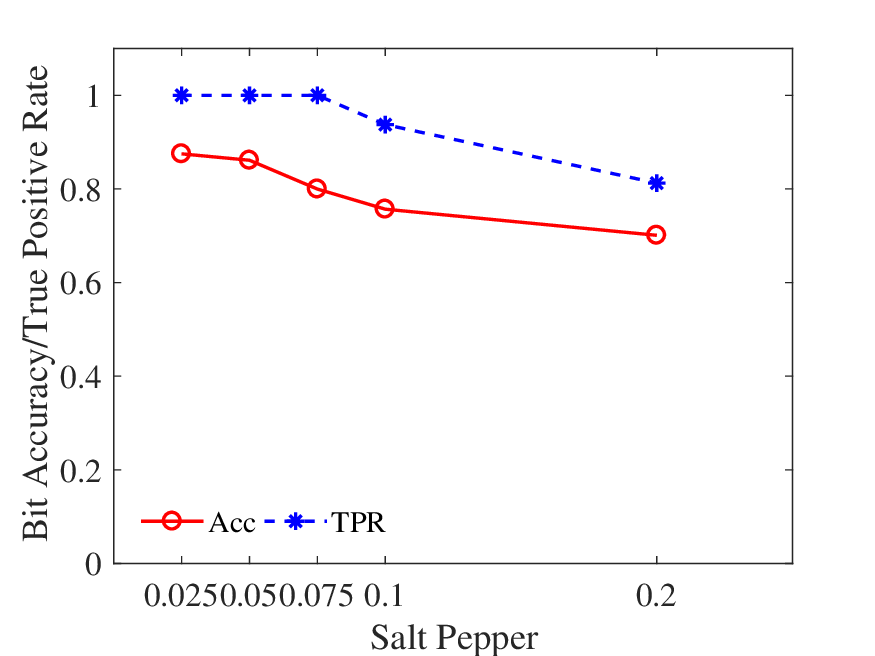} 
    \end{minipage}
}\vspace{-2pt}\hspace{1pt}

\subfloat[Gaussian Blur]
{
    \begin{minipage}[b]{.2\linewidth}
        \centering
        \includegraphics[width=1.7in]{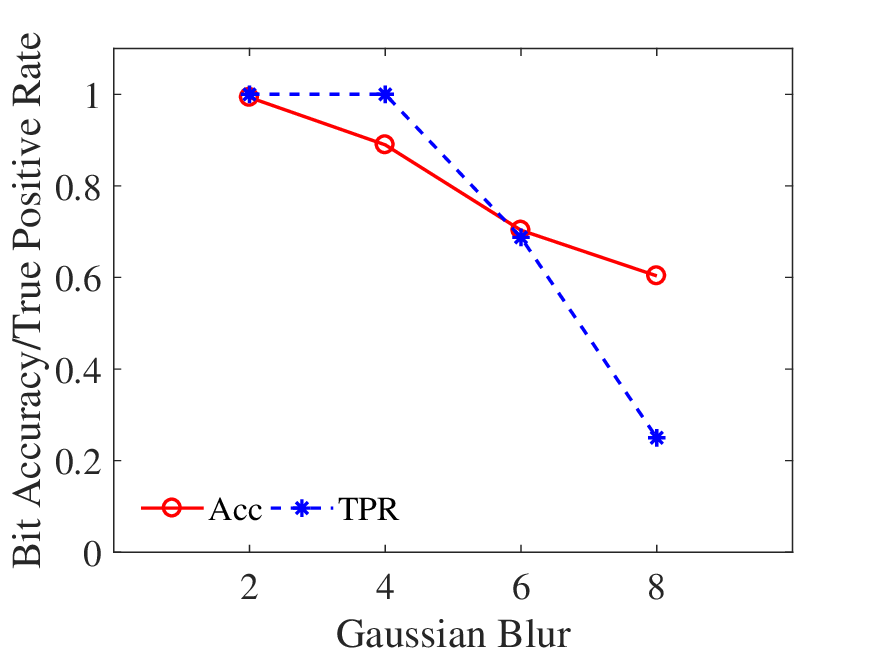}
    \end{minipage}
}\hspace{1pt}
\subfloat[Gaussian Noise]
{
    \begin{minipage}[b]{.2\linewidth}
        \centering
        \includegraphics[width=1.7in]{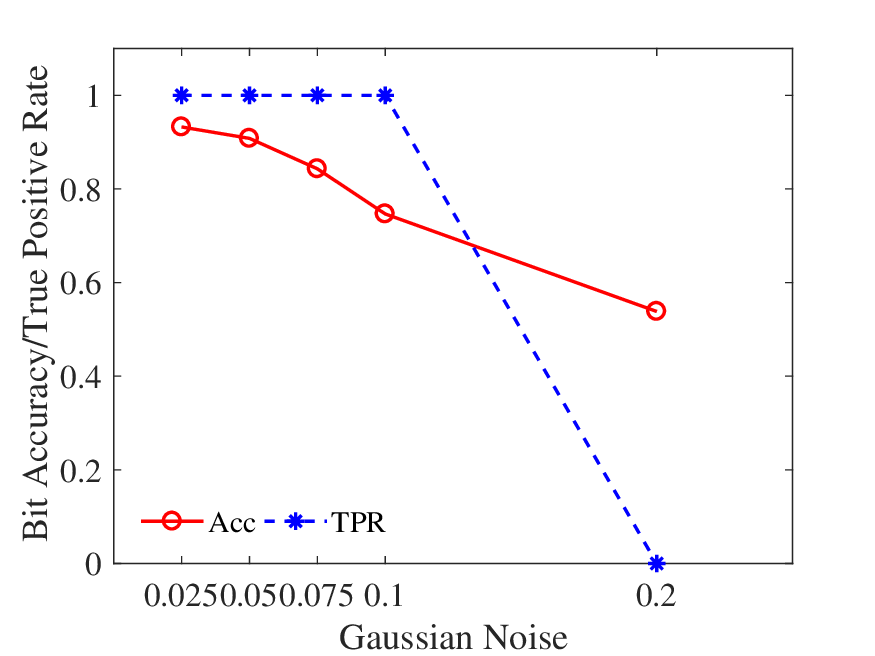} 
    \end{minipage}
}\hspace{1pt}
\subfloat[Random Crop]
{
    \begin{minipage}[b]{.2\linewidth}
        \centering
        \includegraphics[width=1.7in]{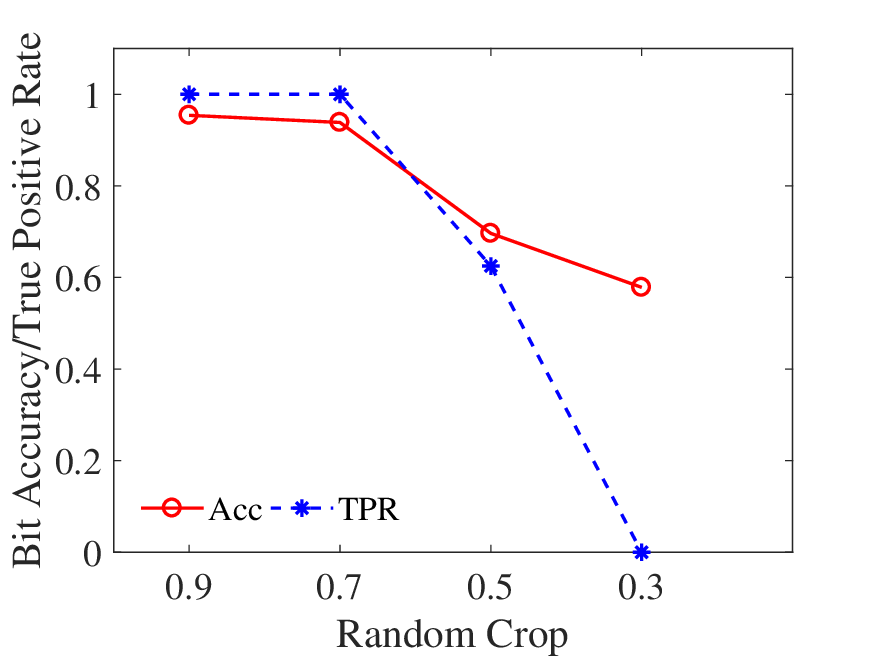}
    \end{minipage}
}\hspace{1pt}
\subfloat[Random Drop]
{
    \begin{minipage}[b]{.2\linewidth}
        \centering
        \includegraphics[width=1.7in]{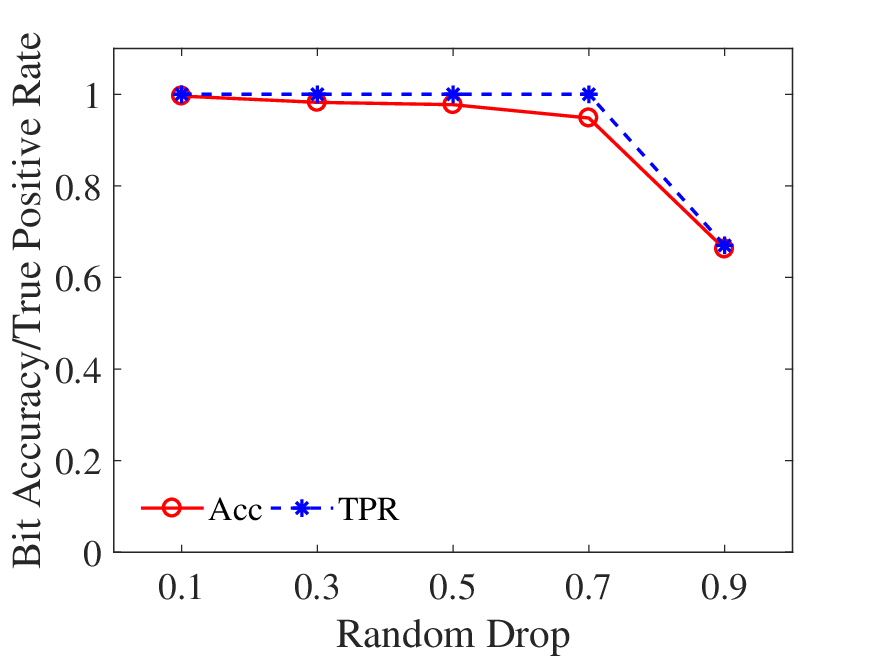} 
    \end{minipage}
}\hspace{1pt}
\caption{TPR and bit accuracy under various image distortions and distortion strengths.}
\label{imagedistortion}
\end{figure*}

{\renewcommand{\arraystretch}{1.2}
\begin{table}[t]
\centering
\caption{Bit accuracy for different Constant Rate Factor (CRF) in H.264.}
\label{TABLE{H264}}
\resizebox{\linewidth}{!}{
\begin{tabular}{c|ccccc}
\hline
\multirow{2}{*}{Methods} & \multicolumn{5}{c}{ CRF (video size compared to CRF=0)}                                 \\ \cline{2-6} 
                         & 18 (25\%)         & 22 (20\%)         & 26 (13.3\%)         & 30 (8.3\%)         & 40 (1.8\%)         \\ \hline
HiDDeN\cite{hidden}                   & 85.43          & 79.85          & 74.64          & 70.08          & 54.67          \\
REVMark\cite{revmark}                  & 99.55          & 95.84          & 93.54          & 85.77          & 68.04          \\
Video Seal\cite{videoseal}               & \textbf{99.94} & 96.03          & 93.11          & 85.56          & 74.65          \\
WAM\cite{WAM}                      & 99.88          & 94.57          & 91.78          & 86.12          & 70.31          \\
DVMark\cite{dvmark}                   & 98.87          & 92.94          & 87.75          & 79.96          & 61.97          \\
ours                     & 99.80          & \textbf{99.02} & \textbf{98.22} & \textbf{96.87} & \textbf{81.05} \\ \hline
\end{tabular}}
\end{table}}

{\renewcommand{\arraystretch}{1.2}
\begin{table}[t]
\centering
\caption{Bit accuracy for different Constant Rate Factor (CRF) in H.265.}
\label{TABLE{H265}}
\resizebox{\linewidth}{!}{
\begin{tabular}{c|ccccc}
\hline
\multirow{2}{*}{Methods} & \multicolumn{5}{c}{CRF (video size compared to CRF=0)}                                 \\ \cline{2-6} 
                         & 18 (25\%)         & 22 (17\%)         & 26 (10\%)         & 30 (6.7\%)         & 40 (1.2\%)         \\ \hline
HiDDeN\cite{hidden}                   & 84.52          & 80.47          & 74.44          & 68.97          & 52.83          \\
REVMark\cite{revmark}                  & 99.56          & 96.36          & 93.12          & 86.48          & 70.09          \\
Video Seal\cite{videoseal}               & \textbf{99.87} & 97.01          & 93.79          & 86.41          & 76.05          \\
WAM\cite{WAM}                      & 99.63          & 93.85          & 90.69          & 85.57          & 69.98          \\
DVMark\cite{dvmark}                   & 98.08          & 90.89          & 85.45          & 77.67          & 61.03          \\
ours                     & 99.32          & \textbf{98.83} & \textbf{96.80} & \textbf{94.14} & \textbf{80.96} \\ \hline
\end{tabular}}
\end{table}}

\subsubsection{\textbf{Image-level Distortion Impact}}
In addition, we also make image attacks on each frame of the video, such as JPEG, Median, Gaussian Blur, etc. All the results are shown in \Cref{imagedistortion}. For Gaussian noise and random clipping, the performance decreases significantly as the strength increases. However, for the other six attacks, even at high intensity, it can still maintains an accuracy rate of greater than 99.9\%.


{\renewcommand{\arraystretch}{1.2}
\begin{table}[t]
\centering
\caption{Bit accuracy with different factors $f_c$, $f_h$ and $f_w$.}
\label{TABLE{fcfh}}
\begin{tabular}{l|ccc}
\hline
$f_c-(f_{h}$=$f_w)$(bits)  & Bit Acc & TPR (detection) & TPR (traceability) \\ \hline
1-2    \ (4096) & 0.7016  & 1.0000            & 1.0000              \\
4-1   \  (4096) & 0.7876  & 1.0000          & 1.0000               \\
1-4   \  (1024) & 0.8898  & 1.0000           & 1.0000               \\
4-2   \  (1024) & 0.8723  & 1.0000            & 1.0000               \\
\textbf{1-8}   \  \textbf{(256)}  & \textbf{1.0000}     & \textbf{1.0000}          & \textbf{1.0000}               \\
4-4  \    (256)  & 0.9980   & 1.0000         & 1.0000               \\
1-16    (64)  & 1.0000    & 1.0000            & 1.0000               \\
4-8  \ (64)   & 1.0000     & 1.0000            & 1.0000               \\ \hline
\end{tabular}
\end{table}}

\subsection{Ablation Studies}
In this section, we perform comprehensive ablation experiments to determine the choice of hyperparameters and modules to validate the effectiveness of our method.

\textbf{Watermark capacity}:We obtained different watermark capacity sizes by adjusting the scaling factors $f_c, f_h$, and $f_w$, and the results are shown in Table \ref{TABLE{fcfh}}. As the watermark capacity increases, while the amount of information that can be embedded is larger, it may also conflict with the prior distribution of the generative model, resulting in reduced extraction accuracy. In the end, we chose $f_c$=1 and $f_h$=$f_w$=8 to get the 256-bit watermark capacity, so we were able to strike a good balance between information and visual quality.

\textbf{Sampling Method}: We evaluated the effectiveness of different continuous time samplers during denoising, as shown in Table \ref{TABLE{6}}. We compare common sampling algorithms, including DDIM, PNDM, DEIS, DPMSolver and UniPC. The results show that DPMSolver has the best performance, mainly due to its high-order numerical integration characteristics, which can significantly reduce the accumulation of discretization errors in the multi-step denoising process. Based on this advantage, we finally chose it.

{\renewcommand{\arraystretch}{1.2}
\begin{table}[t]
\centering
\caption{Bit accuracy with different sampling methods.}
\label{TABLE{6}}
\begin{tabular}{l|ccc}
\hline
Sampling Methods & Bit Acc & TPR (detection) & TPR (traceability) \\ \hline
DDIM \cite{denoising}            & 0.9983  & 1.0000         & 1.0000            \\
PNDM \cite{Pseudonumerical}            & 0.9822  & 1.0000         & 1.0000            \\
DEIS \cite{fastsampling}            & 0.9983  & 1.0000         & 1.0000            \\
\textbf{DPMSolver} \cite{dpmsolver}       & \textbf{0.9992}  & \textbf{1.0000}         & \textbf{1.0000}            \\
UniPC \cite{unipc}           & 0.9989  & 1.0000         & 1.0000            \\ \hline
\end{tabular}
\end{table}}

{\renewcommand{\arraystretch}{1.2}
\begin{table}[t]
\centering
\caption{Bit accuracy of SKeDA with and without DA Module.}
\label{TABLE{weightnet}}
\begin{tabular}{c|cccc}
\hline
 & \begin{tabular}[c]{@{}c@{}}Video   \\ Distortions\end{tabular} & \begin{tabular}[c]{@{}c@{}}Image   \\ Distortions\end{tabular} & \begin{tabular}[c]{@{}c@{}}Average \\   Acc\end{tabular} & \begin{tabular}[c]{@{}c@{}}Average  \\  TPR\end{tabular} \\ \hline
DA Module (w/o) & 0.9431                             & 0.8647                                        & 0.9039                                 & 1.0000                                 \\
\textbf{DA Module (w)}   & \textbf{0.9853}                                    & \textbf{0.9274}                                  & \textbf{0.9564}                             & \textbf{1.0000}                               \\ \hline
\end{tabular}
\end{table}}

\textbf{DA Module}: We evaluated the weight allocator during the extraction process, and its impact is shown in Table \ref{TABLE{weightnet}}. The addition of the DA module significantly improved the model's extraction accuracy in the presence of distorted images and videos. This module can adaptively assign weights according to the correlation of information between frames and quality, emphasizing the retention of more watermark information, so as to realize dynamic weighted fusion in the time dimension. Experiments show that compared with fixed or average weighting strategies, it has a good effect on accurate watermark extraction.

\begin{figure}[t]
    \centering
   \includegraphics[width=2.7in]{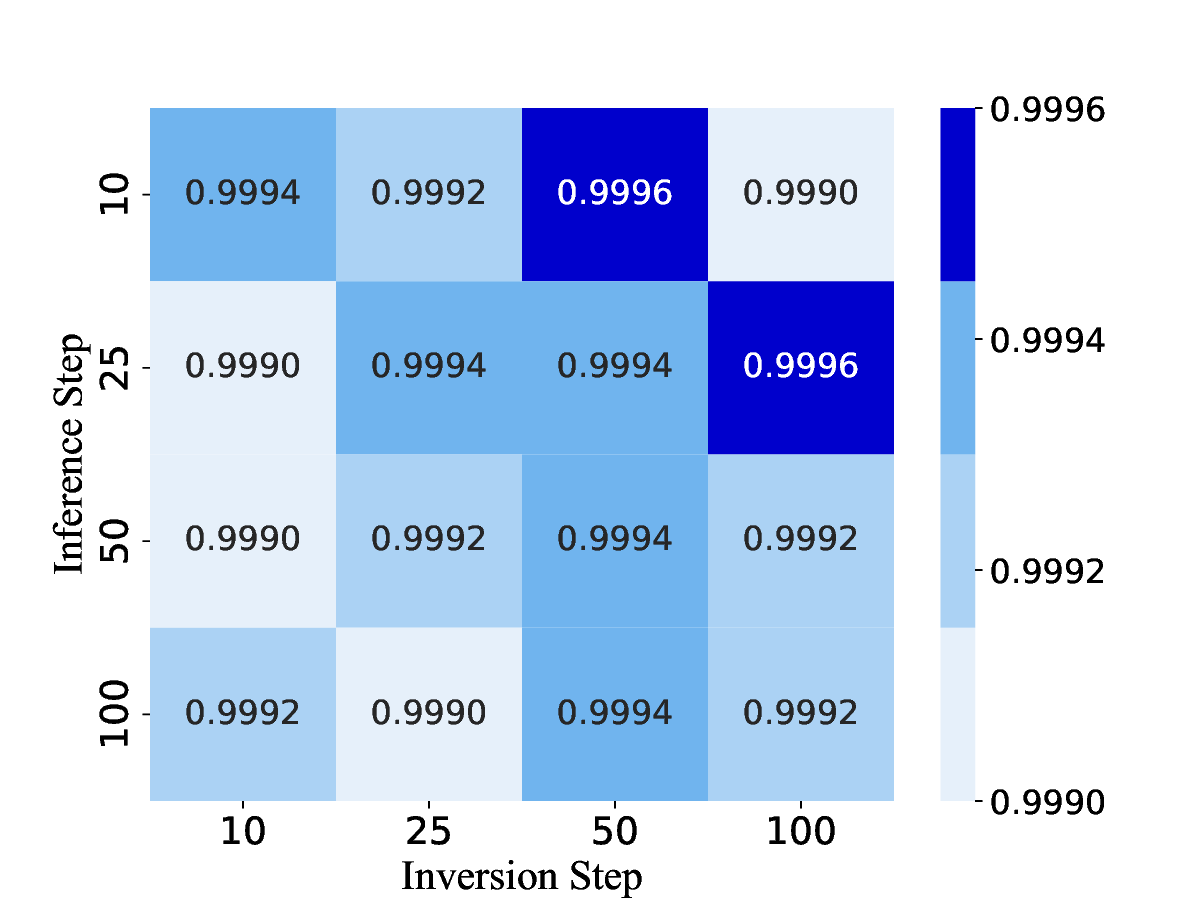}
    \caption{Bit accuracy with different inference and inversion steps.}
    \label{fig:heatmap}
\end{figure}

\textbf{Inversion step}: The settings for the inference and inversion steps may affect the potential variables obtained through DDIM inversion. We vary the inference and inversion steps from 10 to 100, tested in various combinations. The results in \Cref{fig:heatmap} show that our method still achieve an accuracy rate of over 99.9\% with different settings.





\section{Conclusion}
In this paper, we proposed a novel watermarking scheme for text-to-video models to address copyright protection, content authenticity, and misuse prevention in AI-generated content. The method incorporates two key modules: the SKe module for embedding a latent watermark in a concealed and uniformly distributed manner while minimizing key overhead, and the DA module for adaptively assigning extraction weights to individual video frames. By embedding the watermark within the diffusion model's latent space prior to video generation, the watermark becomes inherently fused with the content, enhancing its robustness against various attacks and transformations. Experimental results demonstrate that, compared to existing video watermarking methods, this approach maintains higher visual quality in watermarked videos while ensuring effective watermark extraction. In the future, we aim to extend the method to support larger watermark capacities, longer video sequences, and more complex generation scenarios.
\end{document}